# Virtual Vector Machine for Bayesian Online Classification

**Thomas P. Minka**
Microsoft Research
7 JJ Thomson Avenue
Cambridge, CB3 0FB, UK

**Rongjing Xiang**
Department of CS
Purdue University
West Lafayette, IN 47907

**Yuan (Alan) Qi**
Departments of CS & Statistics
Purdue University
West Lafayette, IN 47907

## Abstract

In a typical online learning scenario, a learner is required to process a large data stream using a small memory buffer. Such a requirement is usually in conflict with a learner's primary pursuit of prediction accuracy. To address this dilemma, we introduce a novel Bayesian online classification algorithm, called the *Virtual Vector Machine*. The virtual vector machine allows you to smoothly trade-off prediction accuracy with memory size. The virtual vector machine summarizes the information contained in the preceding data stream by a Gaussian distribution over the classification weights plus a constant number of virtual data points. The virtual data points are designed to add extra non-Gaussian information about the classification weights. To maintain the constant number of virtual points, the virtual vector machine adds the current real data point into the virtual point set, merges two most similar virtual points into a new virtual point or deletes a virtual point that is far from the decision boundary. The information lost in this process is absorbed into the Gaussian distribution. The extra information provided by the virtual points leads to improved predictive accuracy over previous online classification algorithms.

## 1 Introduction

In an online classification problem, the learning algorithm receives labeled data points in a sequential manner. Using a fixed-size memory buffer, the algorithm is expected to extract information from the data for the purpose of making accurate predictions on test data. A classical example of such an algorithm is the Perceptron algorithm for binary (two-class) classification. For online regression problems, a classical algorithm is the Kalman filter. For linear-Gaussian regression, the Kalman filter is an exact algorithm, in the sense that it retains all of the information in the data necessary to make optimal predictions.

In general, you can construct online learning algorithms by following a Bayesian paradigm (Opper & Winther, 1999). Given a statistical model of the data, you maintain a posterior distribution on the model parameters. As each data point arrives, the posterior distribution is updated. To make predictions, you average according to your uncertainty in the parameters. The Kalman filter can be seen as a special case of this method. To keep the method within a memory bound, you will typically need to approximate the posterior distribution. An efficient and effective approach to this is called assumed-density filtering (ADF) (Opper & Winther, 1999; Minka, 2001). ADF maintains an approximate posterior distribution within a given family $\mathcal{F}$. Upon receiving a new point, the posterior is updated exactly and then projected back onto $\mathcal{F}$ by finding the distribution with minimum Kullback-Leibler divergence. In linear classification problems, the approximating family is typically Gaussian. Gaussian ADF has been used for online training of Gaussian process classifiers and achieved impressive results (Csató, 2002).

In this paper, we describe a novel Bayesian online learning algorithm that extends Gaussian ADF. The algorithm allows you to smoothly improve prediction accuracy as more memory is available. The central idea of this algorithm is to store both a Gaussian distribution and a data cache. The data cache is dynamically updated to hold representative data points from the data stream. In this way, the algorithm maintains both Gaussian and non-Gaussian information about the model parameters. The points in the data cache are *virtual* data points that summarize multiple real points. Therefore, we name this algorithm the Virtual

Vector Machine (VVM). The number of virtual data points is determined by the user.

We discuss related online algorithms, such as the Topmoumoute online natural gradient method (Le Roux et al., 2007), the online Passive-Aggressive algorithm (Crammer et al., 2006), and sparse online Gaussian process classification (Csató & Opper, 2002), in section 4. Section 5 evaluates the VVM and these previous algorithms on both real and synthetic classification problems. We find that the combination of Gaussian and non-Gaussian information in the VVM leads to improved prediction accuracy.

## 2 Online binary classification

From the Bayesian perspective, the task of online learning is to sequentially update the posterior of model parameters based on each incoming new data point and then use the updated posterior distribution to make prediction for the next incoming data point. Specifically, letting $\mathbf{w}$ denote the model parameters, $(\mathbf{x}_1, \ldots, \mathbf{x}_T)$ denote the data from time 1 to time $T$, and $f(x_t; \mathbf{w})$ denote the likelihood function for the data point at time $t$, the posterior at time $T$ is

$$p(\mathbf{w}|\mathbf{x}_1, \ldots, \mathbf{x}_T) \quad (1)$$

$$= \frac{p(\mathbf{w}) \prod_{t=1}^{T} f(x_t; \mathbf{w})}{\int p(\mathbf{w}) \prod_{t=1}^{T} f(x_t; \mathbf{w}) \mathrm{d}\mathbf{w}} \quad (2)$$

where $p(\mathbf{w})$ is the prior distribution on the model parameter $\mathbf{w}$.

The specific problem we consider is online binary classification, using a linear classification function. The likelihood is a step function with labeling error rate $\epsilon \in [0, 1]$:

$$f(\mathbf{x}; \mathbf{w}) = \epsilon(1 - \Theta(\mathbf{w}^\mathrm{T}\mathbf{x})) + (1 - \epsilon)\Theta(\mathbf{w}^\mathrm{T}\mathbf{x}) \quad (3)$$

$$\Theta(z) = \begin{cases} 1 & \text{if } z > 0 \\ 0 & \text{if } z \leq 0 \end{cases} \quad (4)$$

This model expects the inner product $\mathbf{w}^\mathrm{T}\mathbf{x}$ to be positive with probability $1 - \epsilon$. To apply this model to classification, the data point $\mathbf{x}$ should be a feature vector scaled by 1 or $-1$ depending on the label. If the data point belongs to the first class, $\mathbf{x}$ is the feature vector itself; if the data point belongs to the second class, $\mathbf{x}$ is the negative of the feature vector. The prior distribution $p(\mathbf{w})$ is uniform over $\mathbf{w}$, or equivalently for this problem, a standard Gaussian distribution $p(\mathbf{w}) \sim \mathcal{N}(\mathbf{0}, \mathbf{I})$.

For this problem, the posterior distribution for $\mathbf{w}$ is piecewise constant over convex polygonal regions. In three dimensions, the posterior can be nicely visualized with planes cutting a sphere (Herbrich et al., 1999; Minka, 2001). To represent these polygons exactly would require a memory buffer that grows with $T$. In (Opper & Winther, 1999; Minka, 2001) the posterior is approximated by a multivariate Gaussian distribution:

$$q(\mathbf{w}) \sim \mathcal{N}(\mathbf{m}_w, \mathbf{V}_w) \quad (5)$$

Given this approximation, the most likely label for a test point $\mathbf{x}$ is $\text{sign}(\mathbf{m}_w^\mathrm{T}\mathbf{x})$.

In a batch setting, if we want to compute $\mathbf{m}_w$ and $\mathbf{V}_w$ from data, we can apply the expectation propagation (EP) algorithm (Minka, 2001). Expectation propagation (Minka, 2001) is a general algorithm to approximate the exact posterior distribution $p(\mathbf{w}|\mathbf{x}_1, \ldots, \mathbf{x}_T) \propto p(\mathbf{w}) \prod_{t=1}^{T} f(x_t; \mathbf{w})$ by a distribution in an exponential family. The approximation is done by approximating each factor in the expression for the posterior, such that their product gives a good approximation to the full posterior. Thus the approximate posterior would have the form:

$$q_{EP}(\mathbf{w}) = \tilde{f}_0(\mathbf{w}) \prod_{t=1}^{T} \tilde{f}_t(\mathbf{w}) \quad (6)$$

where $\tilde{f}_t(\mathbf{w})$ and $\tilde{f}_0(\mathbf{w})$ are approximations of the likelihood factor $f(x_t; \mathbf{w})$ and the prior $p(\mathbf{w})$, respectively. For the classification problem, the approximate terms are all Gaussian. Since the prior is Gaussian already, we have $\tilde{f}_0(\mathbf{w}) = p(\mathbf{w})$. EP iteratively refines each factor approximation using information from the corresponding exact factor and all the other factor approximations and will loop over all the data points up to the current time for multiple iterations. As a result, EP is impractical for online learning.

However, we can apply restricted forms of EP to do online learning. For example, suppose we process the factors $f(x_t; \mathbf{w})$ in temporal order to get $\tilde{f}_t$, but never refine the previous $\tilde{f}_t$'s. This requires only a fixed amount of memory because we can just store the product of the $\tilde{f}_t$'s in a single Gaussian distribution. In fact, this approach is equivalent to the standard Gaussian ADF algorithm. Another approach is to store a buffer of the last $k$ points and their approximate terms, and only refine these. This leads to window-based EP (Qi & Minka, 2007). We build on these ideas in the next section to obtain the Virtual Vector Machine.

## 3 Virtual Vector Machine

The Virtual Vector Machine is obtained by considering a richer approximating family for the posterior distribution over $\mathbf{w}$ than a Gaussian. We let the approximating distribution be a Gaussian term times a set of virtual data likelihoods:

$$q(\mathbf{w}) \propto r(\mathbf{w}) \prod_i f(\mathbf{b}_i; \mathbf{w}) \qquad (7)$$

$$r(\mathbf{w}) \sim \mathcal{N}(\mathbf{m}_r, \mathbf{V}_r) \qquad (8)$$

where $\mathbf{b}_i$ is the $i$th virtual data point and $f(b_i; \mathbf{w})$ has the form of the original likelihood factors. The Gaussian $r(\mathbf{w})$ is called the "residual"; it represents information from the real data that is not included in the virtual points. Because the likelihood terms $f$ are step functions, the resulting distribution on $\mathbf{w}$ is a Gaussian modulated by a piecewise constant function. If $\epsilon = 0$, then $q(\mathbf{w})$ is a truncated multivariate Gaussian.

From this augmented representation, we can extract a fully Gaussian approximation simply by running EP over the virtual data points with prior $r(\mathbf{w})$. The result will be an approximation to $q(\mathbf{w})$ of the form $\tilde{q}(\mathbf{w}) \sim \mathcal{N}(\mathbf{m}_w, \mathbf{V}_w)$ where

$$\tilde{q}(\mathbf{w}) = r(\mathbf{w}) \prod_i \tilde{f}_i(\mathbf{w}) \qquad (9)$$

and $\tilde{f}_i(\mathbf{w})$ is Gaussian. The Gaussian $\tilde{q}$ is useful as a surrogate for computations on $q$. For example, to classify a test point $\mathbf{x}$, we use $\text{sign}(\mathbf{m}_w^T \mathbf{x})$.

To compute $q(\mathbf{w})$ from a stream of data, we apply Bayesian online learning. That is, at each timestep we have a $q(\mathbf{w})$ computed from the data points until now, and we want to update $q(\mathbf{w})$ in light of the new point. Let $q^{new}(\mathbf{w})$ be the new approximation that we are trying to find. In the spirit of assumed-density filtering, we should try to minimize a distance measure such as:

$$D = \text{KL}(q_+(\mathbf{w}) \parallel q^{new}(\mathbf{w})) \qquad (10)$$

$$\text{where } q_+(\mathbf{w}) = q(\mathbf{w}) f(\mathbf{x}_T; \mathbf{w}) \qquad (11)$$

Note that $q_+$ is also of the form (7), but with one additional data point. So the problem reduces to taking a distribution of the form (7) and approximating it with a distribution of the same form, having one fewer virtual point. An exact minimization of $D$ over the parameters $(\mathbf{B}, \mathbf{m}_r, \mathbf{V}_r)$ would be too costly. Instead we apply a heuristic method to do the reduction. The first step is to run EP on $q_+$ to get a fully Gaussian $\tilde{q}_+$ with approximate factors $\tilde{f}_i$. Then we consider either evicting a virtual point or merging two points, as described below. Each of these possibilities is scored and the one with the best score is chosen to be $q^{new}$.

To score a candidate $q^{new}$, we use an easy-to-compute surrogate for $D$. Intuitively, we want $q^{new}$ to discard the factors which can be well approximated by Gaussians, and keep the non-Gaussian information contained in $q_+$. For example, a sharp truncation near the middle of $r(\mathbf{w})$ should be kept, while a truncation in the tail of $r(\mathbf{w})$ could be discarded. This motivates *maximizing* the non-Gaussianity of $q^{new}$. Non-Gaussianity could be measured by the divergence between $q^{new}$ and $\tilde{q}^{new}$:

$$D_2 = \text{KL}(q^{new}(\mathbf{w}) \parallel \tilde{q}^{new}(\mathbf{w})) \qquad (12)$$

This is difficult to compute exactly, but it can be approximated well by adding up the term-by-term KL divergences:

$$E = \sum_i \text{KL}(f(\mathbf{b}_i; \mathbf{w})\tilde{q}^{new}(\mathbf{w})/\tilde{f}_i(\mathbf{w}) \parallel \tilde{q}^{new}(\mathbf{w})) \qquad (13)$$

See appendix A for details. Because the changes we make to $q_+$ are sparse, we can efficiently compute $E$ for each $q^{new}$ by precomputing all of the divergences from $q_+$ to $\tilde{q}_+$ and then recomputing the divergences only for the factors that changed. Note that this error measure assumes that the proposed $q^{new}$ does not add any new information to $q_+$. We ensure this by considering only two kinds of changes.

The VVM algorithm is summarized in Algorithm 1.

---

**Algorithm 1**: Virtual Vector Machine

1. Initialize $r(\mathbf{w})$ as the prior $p(\mathbf{w})$.
2. Initialize the virtual point set $\mathbf{B}$ to be the first $B$ incoming data points.
3. For each incoming training data point $\mathbf{x}_t$:
   a. Add $\mathbf{x}_t$ into $\mathbf{B}$.
   b. Run EP on $\mathbf{B}$ and $r(\mathbf{w})$ to obtain $\tilde{q}_+(\mathbf{w})$.
   c. Score all possible evictions of virtual points.
   d. Find the $k$ closest pairs of virtual points.
      Test merging each of these pairs.
      For each merge, find the residual $g(\mathbf{w})$
      by solving an inverse ADF projection.

---

### 3.1 Eviction

In the simplest version of the algorithm, to reduce $q_+$ you take one of the likelihoods $f(\mathbf{b}_j; \mathbf{w})$ and replace it with the Gaussian $\tilde{f}_j(\mathbf{w})$ computed earlier by EP. The net effect is to remove $\mathbf{b}_j$ from the cache and to update the residual as follows:

$$r^{new}(\mathbf{w}) = r(\mathbf{w})\tilde{f}_j(\mathbf{w}) \qquad (14)$$

An interesting property of this eviction rule is that if we run EP on $q^{new}$ to get $\tilde{q}^{new}$, there will be a fixed point where all $\tilde{f}_i^{new}(\mathbf{w}) = \tilde{f}_i(\mathbf{w})$. To see this, note that $\tilde{q}^{new}(\mathbf{w}) = r^{new}(\mathbf{w}) \prod_{i \neq j} \tilde{f}_i(\mathbf{w}) =$ $r(\mathbf{w}) \prod_i \tilde{f}_i(\mathbf{w}) = \tilde{q}_+(\mathbf{w})$. When we try to refine $\tilde{f}_i$, we will minimize divergence to the distribution $f(b_i; \mathbf{w})\tilde{q}^{new}(\mathbf{w})/\tilde{f}_i(\mathbf{w})$. But this is the same as the distribution $f(b_i; \mathbf{w})\tilde{q}_+(\mathbf{w})/\tilde{f}_i(\mathbf{w})$ whose divergence was already minimized. As a result of this property, the score $E$ is easy to compute. The divergence for factor $j$ disappears (since it is now Gaussian), and the divergence for other factors is unchanged. Since we want to maximize $E$, this implies that the best point to evict is the one with smallest divergence (the factor which can be best approximated by a Gaussian). It turns out that the divergence for a point $\mathbf{x}$ is a one-dimensional function of its margin: $\mathcal{R}(\mathbf{x}) = \mathbf{m}_w^T \mathbf{x}/\sqrt{\mathbf{x}^T \mathbf{V}_w \mathbf{x}}$. A simple approximation to maximizing $E$ is to pick the point with largest absolute margin: $|\mathcal{R}(\mathbf{x})|$.

### 3.2 Merging

In merging, we replace two non-Gaussian factors $f(\mathbf{b}_1; \mathbf{w})f(\mathbf{b}_2; \mathbf{w})$ with a single non-Gaussian factor and a Gaussian correction: $f(\mathbf{b}'; \mathbf{w})g(\mathbf{w})$. The net effect is to remove $(\mathbf{b}_1, \mathbf{b}_2)$ from the cache, insert $\mathbf{b}'$ into the cache, and to update the residual via $r^{new}(\mathbf{w}) = r(\mathbf{w})g(\mathbf{w})$. The residual term $g$ is important for capturing the lost information from the original two factors. For example, by including $g$, eviction becomes a special case of merging where $\mathbf{b}' = \mathbf{b}_2$, $g = \tilde{f}_1$.

Because merging is an expensive operation, we save computation time by only considering to merge the $k$ pairs of virtual points that are closest together by Euclidean distance (after normalizing each point to unit norm). Here $k$ is a parameter to control computation time. If $k = 1$ then we only consider merging the pair of virtual points that are closest together. Also in the interests of time, we only consider the midpoint $\mathbf{b}' = (\mathbf{b}_1 + \mathbf{b}_2)/2$.

The residual $g$ is computed by approximately minimizing the divergence:

$$D_{12} = \begin{aligned} \text{KL}(f(\mathbf{b}_1; \mathbf{w})f(\mathbf{b}_2; \mathbf{w})\tilde{q}_+^{\backslash 12}(\mathbf{w}) \; || \\ f(\mathbf{b}'; \mathbf{w})g(\mathbf{w})\tilde{q}_+^{\backslash 12}(\mathbf{w})) \end{aligned} \quad (15)$$

$$\text{where } \tilde{q}_+^{\backslash 12}(\mathbf{w}) = \frac{\tilde{q}_+(\mathbf{w})}{\tilde{f}_1(\mathbf{w})\tilde{f}_2(\mathbf{w})} \quad (16)$$

We approximately minimize this by matching the moments of the left and right distributions:

$$\begin{aligned} \text{proj}\Big[f(\mathbf{b}_1; \mathbf{w})f(\mathbf{b}_2; \mathbf{w})\tilde{q}_+^{\backslash 12}(\mathbf{w})\Big] = \\ \text{proj}\Big[f(\mathbf{b}'; \mathbf{w})g(\mathbf{w})\tilde{q}_+^{\backslash 12}(\mathbf{w})\Big] \end{aligned} \quad (17)$$

This leads to two sub-problems:

1. Find the moments of $f(\mathbf{b}_1; \mathbf{w})f(\mathbf{b}_2; \mathbf{w})\tilde{q}_+^{\backslash 12}(\mathbf{w})$.

2. Find $g(\mathbf{w})$ such that $f(\mathbf{b}'; \mathbf{w})g(\mathbf{w})\tilde{q}_+^{\backslash 12}(\mathbf{w})$ has those moments.

For step 1, it is not good enough to use the moments computed by EP (i.e. the moments of $\tilde{q}_+$). In our experience, this does not give an accurate residual and as a result, merging never gets chosen over eviction. Intuitively, the points we would like to merge have $\mathbf{b}_1 \approx \mathbf{b}_2$ and this is precisely the case where EP gives a bad approximation. For the likelihood (3), the moments can be written in terms of the bivariate normal cdf, which we compute exactly using the quadrature method of Genz (2004).

For step 2, we have an inverse problem to the one solved by ADF: we want to find a Gaussian "prior" $g(\mathbf{w})\tilde{q}_+^{\backslash 12}(\mathbf{w})$ such that the "posterior" $f(\mathbf{b}'; \mathbf{w})g(\mathbf{w})\tilde{q}_+^{\backslash 12}(\mathbf{w})$ has given moments. Therefore we call this algorithm *inverse ADF*.

**Inverse ADF Projection**

Denote the unknown prior moments by $(\mathbf{m}_w, \mathbf{V}_w)$ and the given posterior moments by $(\mathbf{m}_w^{new}, \mathbf{V}_w^{new})$. For the likelihood in (3), these are related by: $(\mathbf{x} = \mathbf{b}')$

$$z = \frac{\mathbf{m}_w^T \mathbf{x}}{\sqrt{\mathbf{x}^T \mathbf{V}_w \mathbf{x}}} \quad (18)$$

$$h = \frac{(1-2\epsilon)\mathcal{N}(z;0,1)}{\epsilon + (1-2\epsilon)\Phi(z)} \quad (19)$$

$$\mathbf{m}_w^{new} = \mathbf{m}_w + \frac{h\mathbf{V}_w \mathbf{x}_i}{\sqrt{\mathbf{x}^T \mathbf{V}_w \mathbf{x}}} \quad (20)$$

$$\mathbf{V}_w^{new} = \mathbf{V}_w - (\mathbf{V}_w \mathbf{x})\left(\frac{h(h+z)}{\mathbf{x}^T \mathbf{V}_w \mathbf{x}}\right)(\mathbf{V}_w \mathbf{x})^T \quad (21)$$

Multiplying (20) by $\mathbf{x}^T$, and multiplying (21) by $\mathbf{x}^T$ and $\mathbf{x}$ respectively, we obtain

$$m_h = \mathbf{m}_w^T \mathbf{x} \quad (22)$$
$$v_h = \mathbf{x}^T \mathbf{V}_w \mathbf{x} \quad (23)$$
$$m_h^{new} = m_h + h\sqrt{v_h} \quad (24)$$

$$v_h^{new} = v_h(1 - h(h+z)) \quad (25)$$

Eliminating $m_h$ and $v_h$ leads to:

$$\left(\frac{m_h^{new}}{h+z}\right)^2 = \frac{v_h^{new}}{1 - h(h+z)} \quad (26)$$

Therefore, the moment matching equations are simplified to a one-dimensional nonlinear equation in $z$:

$$\begin{aligned} f(z) = v_h^{new} z^2 + \left(v_h^{new} + (m_h^{new})^2\right) h^2 \\ + \left(2v_h^{new} + (m_h^{new})^2\right) hz - (m_h^{new})^2 = 0 \end{aligned} \quad (27)$$

Its gradient at $z$ is given by

$$f'(z) = 2v_h^{new}z + \left(2v_h^{new} + (m_h^{new})^2\right)h - h(z+h)$$
$$\left(\left(2v_h^{new} + 2(m_h^{new})^2\right)h + \left(2v_h^{new} + (m_h^{new})^2\right)z\right) \quad (28)$$

We then use Levenberg-Marquardt algorithm to find the root $\hat{z}$ of (26) based on (27) and (28). Then $v_h$ and $m_h$ are obtained from

$$v_h = \left(\frac{m_h^{new}}{\hat{z}+h}\right)^2 \quad (29)$$

$$m_h = \hat{z}\sqrt{v_h} \quad (30)$$

and from these we can recover $(\mathbf{m}_w, \mathbf{V}_w)$.

This merging rule has the useful property that $\tilde{q}^{new} = \tilde{q}_+$, just as in eviction. Thus the score $E$ is easy to compute by subtracting the divergence terms for the merged points and adding a divergence term for the new point.

### 3.3 Nonlinear classification via Random Feature Expansion

To apply the virtual vector machine to nonlinear classification tasks, instead of kernelizing them, we adopt the recently developed random feature expansion technique (Rahimi & Recht, 2007). In essence, this technique replaces the kernel trick by mapping the data into a randomized feature space with finite dimension, such that the inner product of two randomly mapped data points is approximately the same as the value of a kernel function evaluated at the two data points. As a result, we obtain a random feature based classifier approximately equivalent to the kernelized classifier.

In the experiments, we use random Fourier features to approximate an RBF Gaussian kernel $k(\mathbf{x} - \mathbf{y}) = e^{\frac{-||\mathbf{x}-\mathbf{y}||^2}{2\sigma}}$. The random Fourier features have the following form:

$$\mathbf{z}(\mathbf{x}) \propto [\cos(\omega_1^T\mathbf{x})\cdots\cos(\omega_D^T\mathbf{x})\sin(\omega_1^T\mathbf{x})\cdots\sin(\omega_D^T\mathbf{x})]^T$$

where $\omega_1, ..., \omega_D$ are iid samples from a special $p(\omega)$. As shown by (Rahimi & Recht, 2007), these random features approximate the kernel $k$:

$$\mathbf{z}(\mathbf{x})'\mathbf{z}(\mathbf{y}) \approx k(\mathbf{x} - \mathbf{y})$$

## 4 Related Work

Many online learning algorithms in the literature are not Bayesian and do not attempt to represent a posterior distribution. They only maintain a current weight vector, and update this vector with each new data point. One design principle for these algorithms is the stochastic gradient method, which started with the Perception algorithm. Another design principle is passive-aggressive learning (Crammer et al., 2006). These approaches are usually very simple to implement and consume very little memory. However they sacrifice performance relative to batch algorithms.

One way to reduce this gap is to also store a covariance matrix. In the stochastic gradient approach, you can update this matrix incrementally, then apply a natural gradient update (Le Roux et al., 2007) or whiten the data prior to a gradient step (Cesa-Bianchi et al., 2005). The covariance matrix can also be updated in a passive-aggressive manner (Crammer et al., 2008), yielding an algorithm similar to ADF.

Virtual vector machines go beyond storing a covariance matrix. Their closest competitor in this respect is window-based EP (Qi & Minka, 2007), which performs EP smoothing based on a sliding window and offers a trade-off between ADF and EP. However, unlike window-based EP that uses the last few points in a data stream, virtual vector machines adaptively choose virtual data points that are representative of past data.

A separate issue is kernelization. When a linear classifier is kernelized, the weight vector (and covariance matrix) are represented implicitly as a linear combination of support vectors. To achieve bounded memory, the number of support vectors must be bounded, which means that the weight vector and covariance matrix are only approximately represented. Online heuristics for choosing a bounded number of support vectors were given by Csató and Opper (2002) and Weston et al. (2005). Note that these support vectors have a different role than the virtual points in the VVM. In this paper, the VVM is not kernelized, so its mean vector and covariance matrix are always represented exactly.

## 5 Experiments

We evaluate the Virtual Vector Machine on both synthetic and real world data for classification tasks and compare its predictive performance with alternative online learning methods.

First, we examine the estimation accuracy of VVM, window-EP and EP for linear classification on synthetic datasets. We sample 150 instances for each class from a mixture of two-dimensional Gaussians. The data instances are almost linearly separable. A bias term is added to the input such that the classifier has three weights. We evaluate each method by comparing the estimated posterior mean after processing the whole data sequence with the exact posterior mean obtained by the Billiard algorithm (Herbrich et al., 1999), a Monte-Carlo method to train Bayesian classifiers. EP iteratively processes the whole dataset multiple times until convergence. The window-EP approach extends EP to online learning by keeping only most recent data points in the buffer and performing EP smoothing on them. For window-EP and VVM, we use the same buffer size and test two buffer sizes, 10 and 40 points. The linear ADF classifier will pass the dataset once without using any buffer. As shown in Figure 1, EP achieves the lowest estimation error, i.e., the highest estimation accuracy. It is not surprising at all since EP iteratively refines the approximation based on all the data points. VVMs provide similar estimation accuracy but with only a buffer size of 40. Furthermore, VVMs outperform Window-EP significantly given the same buffer size. This demonstrates that the virtual points make a huge difference by sensibly summarizing all the history information instead of simply containing information from most recent data points. Without any smoothing on previous information, the ADF classifier obtains the biggest estimation error. Note that VVM can be viewed as a generalization of EP, window-EP and ADF: if we set the buffer size to one, VVM reduces to ADF; if we force all the virtual points to be the last few data points, VVM reduces to window-EP; and if the buffer contains the whole dataset, VVMs is the same as EP. Clearly VVM gives its user the freedom to choose a good trade-off between the memory cost and estimation accuracy.

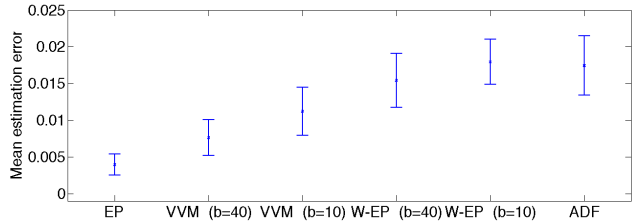

Figure 1: Mean square error of estimated posterior mean obtained by EP, virtual vector machine , ADF and window-EP (W-EP). The exact posterior mean is obtained via a Monte Carlo method. EP performs smoothing on the full dataset of 300 points; VVM and window-EP use the two buffer sizes (10 and 40 points); and ADF effectively use a buffer of size one. The results are averaged over 20 runs.

To further examine the trade-off between the buffer size and the prediction accuracy, we test VVM and window-EP on the UCI Thyroid dataset. The results are summarized in the table 1. When the buffer size increases, the predictive performance of VVM and window-EP improves. Overall, VVM achieves slightly higher accuracy than window-EP, although the advantage is not significant on this dataset.

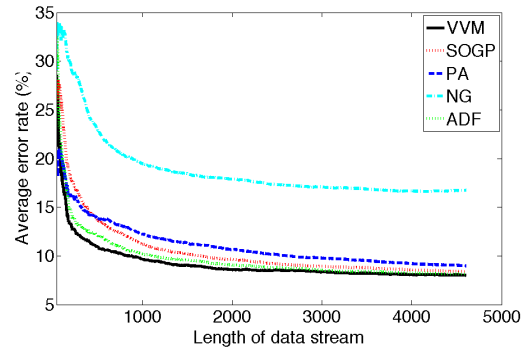

Figure 2: Accumulative prediction error rates of VVM, the sparse online Gaussian process classifier (SOGP), the Passive-Aggressive (PA) algorithm and the Topmoumoute online natural gradient (NG) algorithm on the *Spambase* dataset. The dataset is randomly permutated 10 times and the results are averaged. The size of virtual point set used by VVM is 30, while the online Gaussian process model has 143 basis points.

On real data, we compare VVMs with ADF and three state-of-art online learning algorithms, including the sparse online Gaussian process algorithm (SOGP) (Csató & Opper, 2002), the online passive-aggressive algorithm (PA) (Crammer et al., 2006), and the Topmoumoute online natural gradient algorithm (NG) (Le Roux et al., 2007). The sparse online Gaussian process algorithm always prunes a basis point when incorporating a new data point. The kernelized PA updates the classifier by deciding whether to include the new data point as a support vector each time. It does not control the growth of the set by deleting points that are no longer "support vectors". Thus in its kernelized form, it is not strictly an online algorithm by our definition. As a result, instead of using nonlinear kernels, we use the Fourier-Gaussian random feature expansion described in section 5 to approximate the nonlinear feature mapping for the PA updates. Since VVM uses the same random feature expansion for nonlinear classifications, we have a more consistent comparison between the passive-aggressive algorithm and VVM by eliminating the difference in the feature input. The natural gradient algorithm updates the classifier each time based on a window of the same size as the buffer size used by VVM.

Table 1: Test error rates of VVM and window-EP on the Thyroid dataset with various buffer sizes. Results are averaged over 10 random permutations of the data. For VVM and window-EP, we use the same random Fourier-Gaussian feature expansion with dimension 100.

| Buffer Size | Window-EP error rate (%) | VVM error rate (%) |
|---|---|---|
| 10 | $8.20 \pm 1.01$ | $\mathbf{7.86 \pm 1.03}$ |
| 20 | $6.64 \pm 0.70$ | $\mathbf{6.60 \pm 0.67}$ |
| 30 | $6.05 \pm 0.97$ | $\mathbf{5.94 \pm 0.91}$ |
| 50 | $\mathbf{4.07 \pm 0.62}$ | $4.15 \pm 0.61$ |

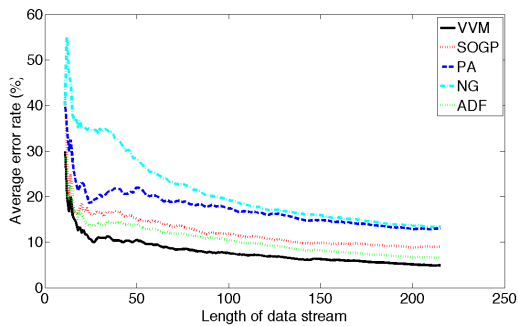

Figure 3: Accumulative prediction error rates of VVM, the sparse online Gaussian process classifier (SOGP), the Passive-Aggressive (PA) algorithm and the Top-moumoute online natural gradient (NG) algorithm on the *Thyroid* dataset. The dataset is randomly permutated 10 times and the results are averaged. VVM, PA, and NG use the same random Fourier-Gaussian feature expansion (dimension 100). NG and VVM both use a buffer to cache 10 points, while the online Gaussian process model and the Passive-Aggressive algorithm have 12 and 91 basis points, respectively.

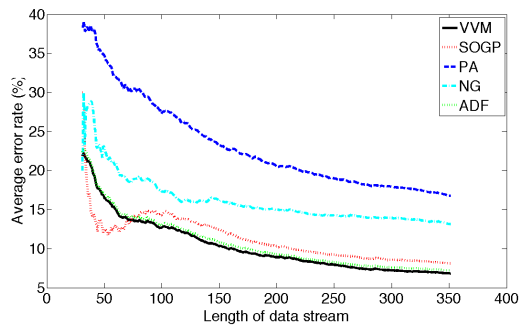

Figure 4: Accumulative prediction error rates of VVM, the sparse online Gaussian process classifier (SOGP), the Passive-Aggressive (PA) algorithm and the Top-moumoute online natural gradient (NG) algorithm on the *Ionosphere* dataset. VVM, PA, and NG use the same random Fourier-Gaussian feature expansion (dimension 100). NG and VVM both use a buffer to cache 30 points, while the online Gaussian process model and the Passive-Aggressive algorithm have 279 and 189 basis points, respectively.

We test the five online learning algorithms on three UCI datasets: Spambase, Thyroid, and Ionosphere. For the online Gaussian process classification, we use a RBF kernel for all the datasets and tune the kernel-width manually to achieve good prediction accuracy. As mentioned before, we use the same random feature expansion for the passive-aggressive algorithm and the VVM algorithm To choose the frequency parameter for the random feature expansion, we first run EP with various frequency parameters and select the one with the smallest validation error. The regularization coefficient used by the passive-aggressive algorithm is 2 and VVM uses a Gaussian prior with mean **0** and variance **I** (the identity matrix).

We evaluate these methods using the average error rate along the data stream. For each new incoming data point, we first make prediction based on the current classifier, compute the error rate up to this point, and then update the classifier using this data point. Finally, the error rates are averaged over 10 runs, where for each run, the dataset is randomly permutated. The experimental results are shown in Figure 2, 3 and 4. They demonstrate the superior prediction accuracy of VVMs.

## 6 Conclusions

In this paper, we have presented a novel Bayesian online learning approach, the Virtual Vector Machine. Built upon two well-known Bayesian learning methods, EP and ADF, the VVM algorithm performs efficient online learning with a small constant space cost regardless of the increasing length of a data stream. We have demonstrated that VVM can achieve improved prediction accuracy on several benchmark datasets.

**Acknowledgement:** We thank the anonymous reviewers for in-depth reviews and great suggestions.

## A    Estimating the error of EP

$$E = \sum_i \mathrm{KL}(\hat{p}_i(\mathbf{w}) \,||\, q(\mathbf{w})) \qquad (31)$$

$$\text{where } \hat{p}_i(\mathbf{w}) \propto f_i(\mathbf{w})p(\mathbf{w})\prod_{j \neq i}\tilde{f}_j(\mathbf{w}) \qquad (32)$$

$$= \frac{1}{Z_i}f_i(\mathbf{w})q^{\backslash i}(\mathbf{w}) \qquad (33)$$

$$q^{\backslash i}(\mathbf{w}) \propto p(\mathbf{w})\prod_{j \neq i}\tilde{f}_j(\mathbf{w}) \qquad (34)$$

$$Z_i = \int_{\mathbf{w}} f_i(\mathbf{w})q^{\backslash i}(\mathbf{w})d\mathbf{w} \qquad (35)$$

At convergence of EP, $\hat{p}_i$ and $q$ have the same expectations. Thus the KL-divergence between $\hat{p}_i$ and $q$ simplifies as follows:

$$\mathrm{KL}(\hat{p}_i(\mathbf{w}) \,||\, q(\mathbf{w})) = \int_{\mathbf{w}} \hat{p}_i(\mathbf{w})\log\frac{\hat{p}_i(\mathbf{w})}{q(\mathbf{w})}d\mathbf{w} \qquad (36)$$

$$= \int_{\mathbf{w}} \hat{p}_i(\mathbf{w})\log\frac{f_i(\mathbf{w})q^{\backslash i}(\mathbf{w})}{Z_i q(\mathbf{w})}d\mathbf{w} \qquad (37)$$

$$= \int_{\mathbf{w}} \hat{p}_i(\mathbf{w})\log f_i(\mathbf{w})d\mathbf{w} + \int_{\mathbf{w}} q(\mathbf{w})\log\frac{q^{\backslash i}(\mathbf{w})}{Z_i q(\mathbf{w})}d\mathbf{w} \qquad (38)$$

where the last line comes from the moment matching property of $\hat{p}_i$ and $q$. Only the first term of (38) depends on the original factor $f_i$.

For example, suppose

$$f_i(\mathbf{w}) = \epsilon(1 - \Theta(\mathbf{w}^{\mathrm{T}}\mathbf{x} - t)) + (1-\epsilon)\Theta(\mathbf{w}^{\mathrm{T}}\mathbf{x} - t) \qquad (39)$$

$$q^{\backslash i}(\mathbf{w}) = \mathcal{N}(\mathbf{w}; \mathbf{m}^{\backslash i}, \mathbf{V}^{\backslash i}) \qquad (40)$$

Then the relevant integral reduces to a 1-D problem:

$$\int_{\mathbf{w}} \hat{p}_i(\mathbf{w})\log f_i(\mathbf{w})d\mathbf{w} = \frac{1}{Z_i}\int_{-\infty}^{\infty} f_i(u)\mathcal{N}(u; m_u, v_u)\log f_i(u)du \qquad (41)$$

$$\text{where } f_i(u) = \epsilon(1-\Theta(u)) + (1-\epsilon)\Theta(u) \qquad (42)$$

$$m_u = \mathbf{x}^{\mathrm{T}}\mathbf{m}^{\backslash i} - t \qquad (43)$$

$$v_u = \mathbf{x}^{\mathrm{T}}\mathbf{V}^{\backslash i}\mathbf{x} \qquad (44)$$

$$E_i = \int_{-\infty}^{\infty} f_i(u)\mathcal{N}(u; m_u, v_u)\log f_i(u)du \qquad (45)$$

$$= E_i^- + E_i^+ \qquad (46)$$

$$\text{where } E_i^- = \int_{-\infty}^{0} f_i(u)\mathcal{N}(u; m_u, v_u)\log f_i(u)du \qquad (47)$$

$$= (\epsilon\log\epsilon)\phi\left(\frac{-m_u}{\sqrt{v_u}}\right) \qquad (48)$$

$$E_i^+ = \int_{0}^{\infty} f_i(u)\mathcal{N}(u; m_u, v_u)\log f_i(u)du \qquad (49)$$

$$= ((1-\epsilon)\log(1-\epsilon))\phi\left(\frac{m_u}{\sqrt{v_u}}\right) \qquad (50)$$

We recognize $\phi\left(\frac{m_u}{\sqrt{v_u}}\right)$ as the leave-one-out predictive probability of $x_i$'s label. Thus the heuristic of leave-one-out predictions arise naturally from considering the KL-divergence of EP. Also note that $E_i \to 0$ as $\epsilon \to 0$.